\documentclass[journal,twoside,web]{ieeecolor}
\usepackage[T1]{fontenc}
\usepackage{jsen}
\usepackage{cite}
\usepackage{amsmath,amssymb,amsfonts}
\usepackage{graphicx}
\usepackage{textcomp}
\usepackage{booktabs}
\usepackage{array}
\usepackage{float}
\usepackage{url}

\newcolumntype{L}[1]{>{\raggedright\arraybackslash}p{#1}}

\def\BibTeX{{\rm B\kern-.05em{\sc i\kern-.025em b}\kern-.08em
    T\kern-.1667em\lower.7ex\hbox{E}\kern-.125emX}}
\definecolor{abstractbg}{rgb}{0.89804,0.94510,0.83137}
\begin{document}
\title{Precision-Aware Illumination-Disentangled Vision Transformer for Spacecraft 6D Pose Estimation}
\author{\centerline{\fontsize{13}{15}\selectfont Zongwu Xie, Yifan Yang, Yonglong Zhang, Guanghu Xie, Yang Liu\textsuperscript{*}, and Shuo Zhang%
\thanks{The authors are with the School of Mechatronics Engineering, Harbin Institute of Technology, Harbin 150001, China (e-mails: xiezongwu@hit.edu.cn; yangyifan03001@163.com; zhangyl202601@163.com; 23b308003@stu.hit.edu.cn; liuyanghit@hit.edu.cn; 15002602293@163.com). \textsuperscript{*}Corresponding author: Yang Liu (e-mail: liuyanghit@hit.edu.cn).}}}

\IEEEtitleabstractindextext{%
\fcolorbox{abstractbg}{abstractbg}{%
\begin{minipage}{\textwidth}%
{\sffamily\bfseries
\noindent\begin{minipage}[t]{0.50\textwidth}
\vspace{0pt}%
\textcolor{subsectioncolor}{\textit{Abstract}}---Vision sensors provide a lightweight solution for spacecraft proximity operations, but monocular spacecraft 6D pose estimation remains difficult under illumination variation, specular reflection, shadowing, weak texture, and background interference. These factors make local visual evidence spatially unreliable and can destabilize pose regression. This article proposes a Precision-Aware Illumination-Disentangled Vision Transformer (PAID-ViT) for robust spacecraft pose estimation.
\end{minipage}\hfill
\begin{minipage}[t]{0.48\textwidth}
\vspace{0pt}%
\centering\includegraphics[width=0.88\linewidth]{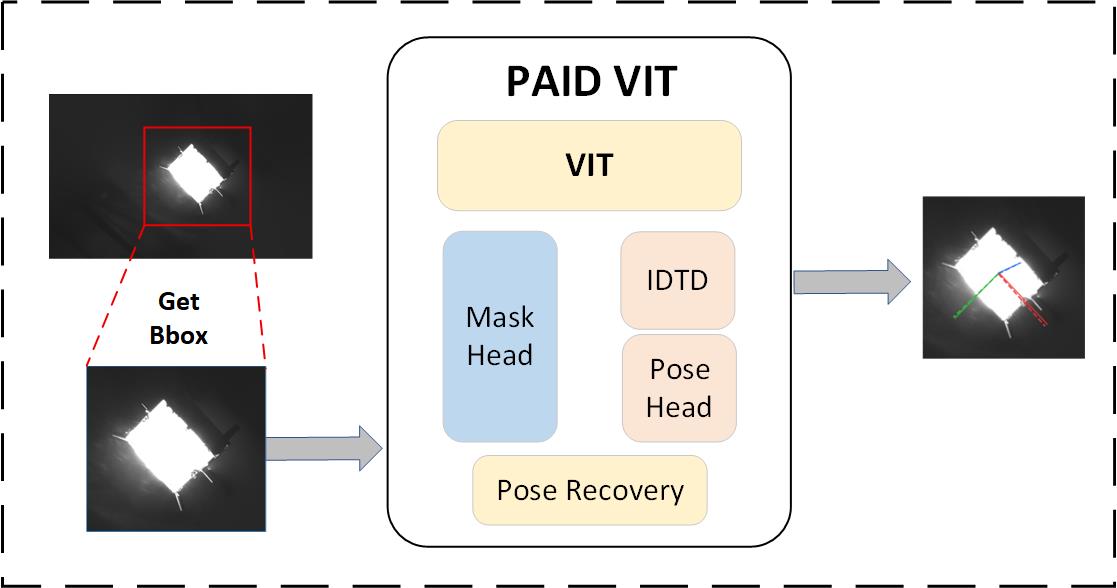}
\end{minipage}
\par\vspace{0.45\baselineskip}\noindent The proposed model separates pose-relevant structure tokens from illumination-sensitive appearance tokens, estimates patch reliability before pose aggregation, and uses foreground mask supervision to preserve silhouette cues. A parameter-free geometric recovery module converts normalized crop coordinates, log-depth, and a continuous 6D rotation representation into camera-frame rotation and translation. Experiments on SPEED+ V2, the SPEED+ validation/lightbox/sunlamp evaluation configuration used in this study, suggest that PAID-ViT reduces translation error and improves robustness in the challenging sunlamp domain, while ablation studies support the complementary roles of illumination disentanglement, reliability-aware token aggregation, mask supervision, and training-side regularization.
\par}

\end{minipage}}}

\IEEEaftertitletext{\vspace{-2.3\baselineskip}}
\maketitle

\section{Introduction}
\label{sec:introduction}
\IEEEPARstart{W}{ith} the rapid development of space technology, modern society increasingly depends on orbital infrastructure for communication, navigation, environmental monitoring, resource investigation, and deep-space exploration. The growth of large-scale low Earth orbit constellations, together with satellite failures, spacecraft retirement, and orbital debris, is making near-Earth space more crowded and hazardous \cite{fcc2022starlinkgen2,esa2026spaceenvironment}. In this context, on-orbit servicing, target capture, in-orbit maintenance, and failed-spacecraft removal require reliable relative pose estimation for autonomous approach, rendezvous, docking, and robotic manipulation.

\begin{figure}[!t]
\centerline{\includegraphics[width=\columnwidth]{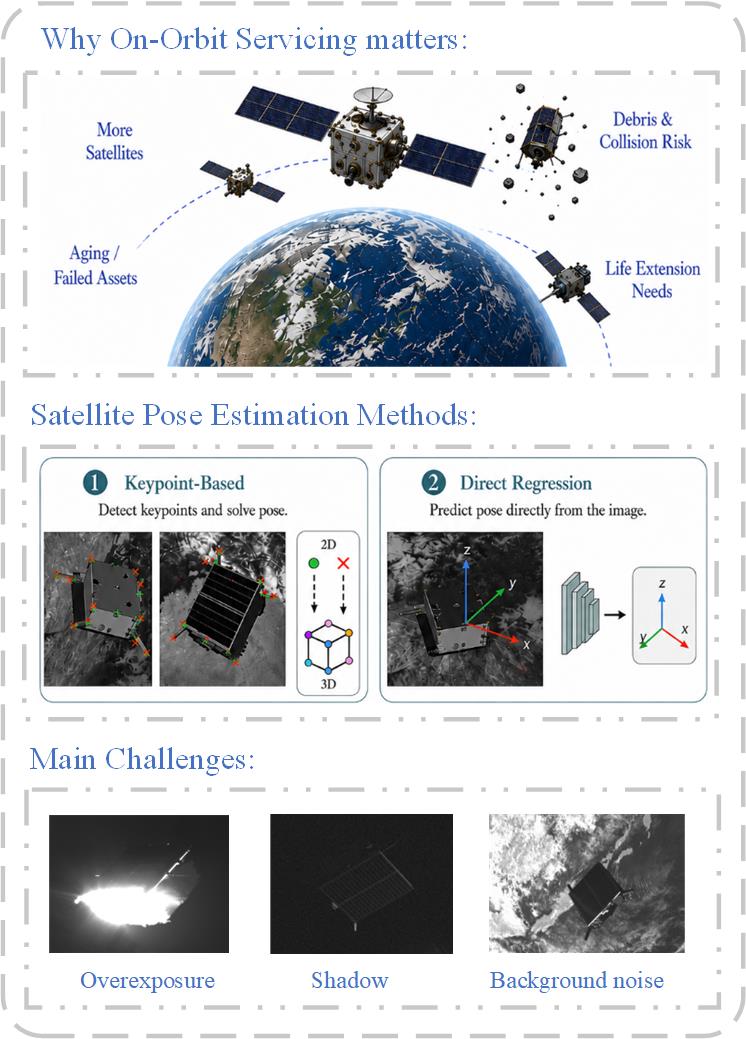}}
\caption{Motivation of spacecraft relative pose estimation for on-orbit servicing and non-cooperative space target perception.}
\label{fig:introduction_motivation}
\end{figure}

Compared with active sensors, monocular cameras are attractive for spaceborne platforms because they are compact, power-efficient, and easy to integrate with optical navigation systems. However, monocular spacecraft 6D pose estimation remains difficult because spacecraft images often contain weakly textured metallic surfaces, repetitive solar-panel structures, thin appendages, self-occlusion, and large appearance variations caused by orbital illumination. These factors make local image evidence highly nonuniform and have been repeatedly identified as major obstacles in spacecraft pose estimation studies and surveys \cite{pauly2023survey,park2022speedplus}.

Early vision-based methods mainly relied on model-based alignment and hand-crafted geometric correspondences, while recent work has moved toward data-driven pose estimation. Sharma \emph{et al.} proposed a robust model-based monocular initialization pipeline for non-cooperative rendezvous \cite{sharma2018modelbased}, and Spacecraft Pose Network later demonstrated neural pose prediction for spacecraft imagery \cite{sharma2019spn}. SPEED+ further introduced synthetic, lightbox, and sunlamp imagery, enabling controlled evaluation of illumination variation and domain shift \cite{park2022speedplus}. Recent satellite Sim2Real data construction also highlights the need to preserve spacecraft component structure while adapting visual appearance \cite{xie2026componentawarestructurepreservingstyletransfer}. In broader robotic perception, PSVMLP and DexMGNet similarly indicate that local geometric structure and spatial aggregation are important for robust 3D understanding and manipulation \cite{xie2024psvmlp,xie2025dexmgnet}.

The remaining challenge is that not all visual regions contribute equally to pose estimation. Spacecraft body edges, panel layouts, and silhouettes usually provide stable geometric evidence, whereas saturated highlights, cast shadows, specular reflections, and background clutter may introduce misleading local tokens. PAID-ViT therefore uses a Vision Transformer backbone to extract global and patch-level tokens, decomposes them through Intrinsic Disentangled Token Decomposition (IDTD), and estimates patch-level reliability so that unreliable regions contribute less to the final pose descriptor. A coarse mask branch provides foreground-aware structural supervision, and a parameter-free recovery module converts intermediate pose variables into camera-frame rotation and translation.

The main contributions of this article are summarized as follows.
\begin{itemize}
\item Method design: a pose-oriented IDTD module separates geometry-relevant structure tokens from illumination-sensitive appearance tokens, and Precision-Aware Uncertainty Pooling (PAUP) aggregates structural patches according to predicted reliability.
\item Training and geometric design: the framework combines coarse mask supervision, illumination prediction, token-orthogonality regularization, training-side patch perturbation, and parameter-free geometric recovery.
\item Experimental evaluation: SPEED+ V2 comparisons and ablations suggest that the proposed components improve direct no-PnP pose regression, especially in translation error and sunlamp-domain robustness.
\end{itemize}

The remainder of this article is organized as follows. Section~\ref{sec:related} reviews related studies on spacecraft pose estimation, generic 6D object pose estimation, visual transformers, and uncertainty-aware learning. Section~\ref{sec:method} presents the proposed PAID-ViT architecture, its major modules, and the training objective. Section~\ref{sec:experiments} gives the experimental design and ablation protocol, and Section~\ref{sec:conclusion} concludes this article.

\section{Related Work}
\label{sec:related}

\subsection{Spacecraft 6D Pose Estimation}
\label{subsec:spacecraft_pose}
Monocular spacecraft pose estimation has evolved from model-based geometric pipelines to learning-based visual regression. Model-based methods explicitly use spacecraft geometry and camera projection constraints, which provides interpretability but may be sensitive to initialization, occlusion, and weak texture \cite{sharma2018modelbased}. Learning-based methods alleviate part of this difficulty by extracting discriminative visual features directly from images. For example, the Spacecraft Pose Network line of work showed that deep networks can provide useful pose predictions for non-cooperative spacecraft imagery \cite{sharma2019spn}.

Benchmark datasets have made it possible to evaluate spacecraft pose estimation under controlled domain gaps. SPEED+ is particularly important because it includes synthetic, lightbox, and sunlamp subsets, thereby exposing the impact of illumination mismatch on learned pose estimators \cite{park2022speedplus}. Recent surveys further emphasize that illumination variation, specular reflection, lack of real annotations, and domain shift remain open problems for spacecraft pose perception \cite{pauly2023survey}. PAID-ViT is designed specifically around these issues by decoupling illumination-sensitive factors and by reducing the influence of unreliable local evidence.

\subsection{Generic 6D Object Pose Estimation}
\label{subsec:generic_pose}
Generic 6D object pose estimation provides useful design principles for spacecraft pose perception. PoseCNN predicts object labels, translation, and rotation from RGB images, demonstrating that deep visual features can support pose estimation in cluttered environments \cite{xiang2018posecnn}. PVNet predicts pixel-wise vectors toward object landmarks and then solves object pose using geometric correspondences, showing the advantage of intermediate geometric representation under occlusion \cite{peng2019pvnet}. Although spacecraft differ from ordinary objects in appearance and operational environment, these studies suggest that pose estimation benefits from structured intermediate variables and auxiliary spatial supervision.

\subsection{Transformers and Uncertainty-Aware Visual Learning}
\label{subsec:vit_uncertainty}
Vision Transformers represent an image as a sequence of patch tokens and a global class token, allowing long-range interaction among spatial regions \cite{dosovitskiy2021vit}. DETR further demonstrated that transformer attention can support structured visual prediction through learned query interactions \cite{carion2020detr}. For spacecraft pose estimation, the tokenized representation is attractive because global tokens can summarize object-level configuration, whereas patch tokens retain local evidence from panels, body contours, and shadowed regions.

Uncertainty-aware learning is also relevant because local image evidence may be noisy or ambiguous. Kendall and Gal introduced uncertainty modeling for deep visual learning, and subsequent work used uncertainty to balance multi-task scene-geometry losses \cite{kendall2017uncertainties,kendall2018multitask}. In PAID-ViT, uncertainty is used at the patch level inside IDTD. Instead of only predicting confidence for the final output, the model estimates the reliability of decomposed structural patches and uses it to adjust attention weights during structure aggregation. Coarse mask supervision is adopted as another auxiliary cue, inspired by segmentation losses such as Dice loss that encourage overlap between predicted and target foreground regions \cite{milletari2016vnet}.

\section{Method}
\label{sec:method}

For spacecraft 6D pose estimation, the key lies in recovering a geometrically consistent camera-frame rotation and translation from image observations that may contain illumination variation, specular reflection, shadowing, and background contamination. This requires the model not only to distinguish pose-relevant spacecraft structures from the background, but also to perceive the reliability of local visual tokens. Given a cropped RGB image $I\in\mathbb{R}^{B\times3\times224\times224}$, PAID-ViT predicts normalized crop coordinates, log-depth, and a continuous 6D rotation representation \cite{zhou2019rotation}, and then recovers the final camera-frame pose using camera geometry. In this section, we first introduce the overall architecture in Section~\ref{subsec:overall}. Section~\ref{subsec:vit_backbone} presents the ViT backbone. Section~\ref{subsec:idtd} presents IDTD, which separates structure and illumination tokens; PAUP, which weights patches according to predicted reliability; and SPM, which perturbs patch reliability during training. Section~\ref{subsec:pose_head} describes the intermediate pose head. Sections~\ref{subsec:mask_branch} and~\ref{subsec:recovery} introduce the coarse mask branch and geometric recovery module, respectively. Section~\ref{sec:loss} gives the training objective of the proposed network.

\subsection{Overall Architecture}
\label{subsec:overall}
\begin{figure*}[!t]
\centerline{\includegraphics[width=0.9\textwidth]{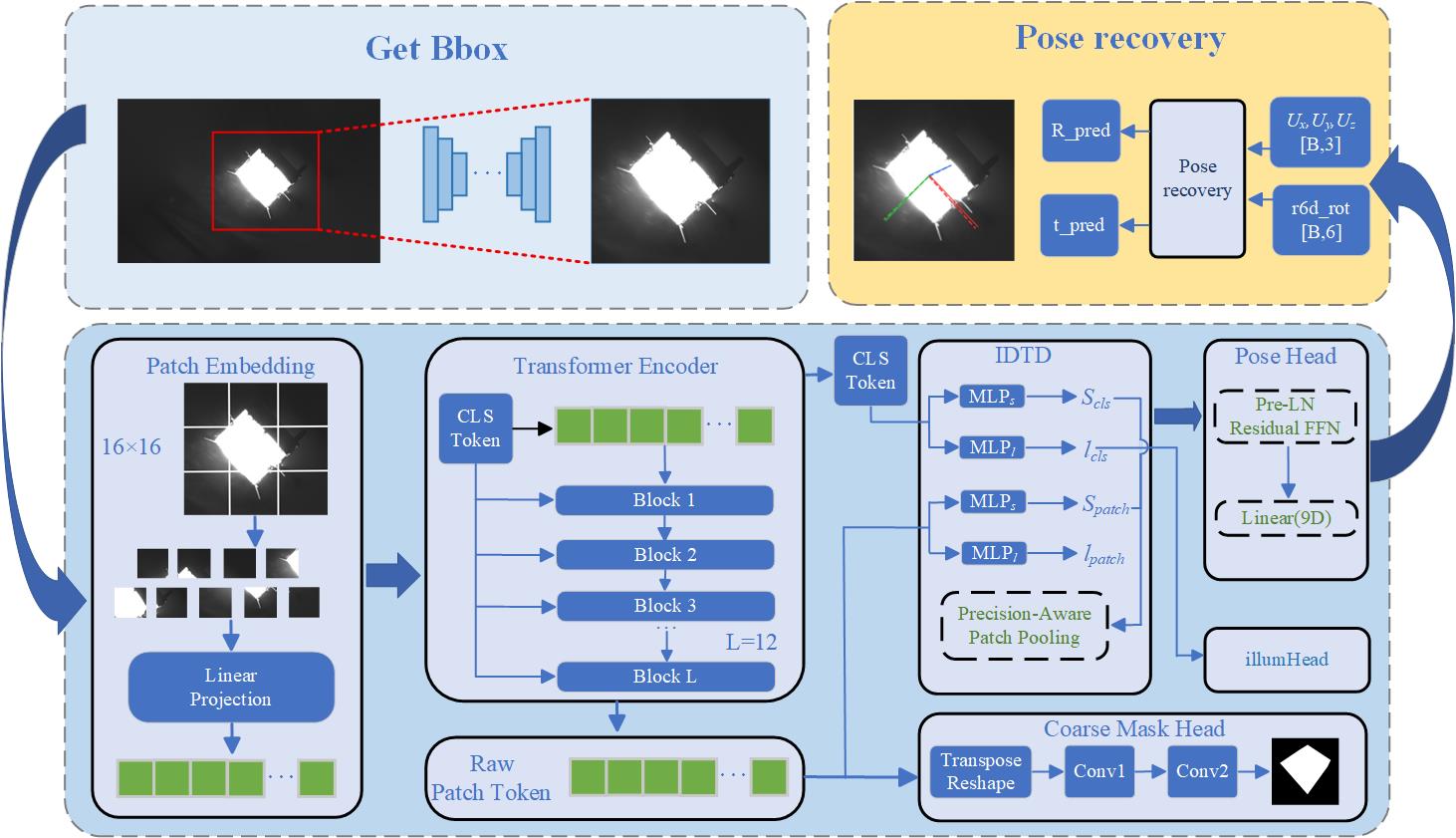}}
\caption{Overall architecture of the proposed PAID-ViT framework.}
\label{fig:overall_architecture}
\end{figure*}

The overall architecture of PAID-ViT is shown in Fig.~\ref{fig:overall_architecture}. The network consists of a ViT backbone, a precision-aware IDTD module, an intermediate pose head, a coarse mask branch, an illumination prediction head, and a parameter-free pose recovery module. The ViT backbone first maps the input image into a global CLS token and a sequence of patch tokens:
\begin{equation}
\mathbf{f}_{cls} \in \mathbb{R}^{B \times C}, \qquad
\mathbf{F}_{patch} \in \mathbb{R}^{B \times N \times C},
\label{eq:backbone_output}
\end{equation}
where $C=768$ and $N=196$ for a ViT-Base encoder with $16\times16$ patches and $224\times224$ input resolution. The CLS token provides global configuration information, whereas the patch tokens retain local evidence from spacecraft components and surrounding background.

Following the feature extraction stage, IDTD decomposes tokens into structure and illumination streams and performs precision-aware aggregation inside the structure stream. The pose head consumes the IDTD output descriptor rather than raw transformer tokens. The mask branch remains attached to raw patch tokens to preserve dense foreground and boundary information. Finally, the recovery module converts the predicted intermediate variables into a physically valid pose. During inference, the final pose is produced by the ViT backbone, IDTD with PAUP, the pose head, and geometric recovery. The mask branch, illumination head, SPM, and consistency losses are used as training-side auxiliary or regularization components and are not required to compute the final pose. This design follows a stage-by-stage processing paradigm: representation extraction, precision-aware intrinsic decomposition, compact intermediate regression, auxiliary foreground supervision, and geometric reconstruction.

\subsection{ViT Backbone}
\label{subsec:vit_backbone}
PAID-ViT adopts a Vision Transformer backbone because spacecraft pose estimation requires both global configuration reasoning and local structural evidence. CNN backbones are effective for local texture extraction, but the visible geometry of a spacecraft is often distributed across distant components, such as the main body, solar panels, antennas, and silhouette boundaries. Self-attention allows patch tokens from these spatially separated regions to interact directly, making ViT suitable for modeling long-range geometric relationships under partial occlusion and illumination variation.

The backbone follows the ViT-Base/16 configuration. The cropped RGB image is resized to $224\times224$ and divided into nonoverlapping $16\times16$ patches, resulting in $N=14\times14=196$ patch tokens. Each patch is linearly projected to a $C=768$ dimensional embedding, and a learnable CLS token is appended to summarize the global object configuration. The transformer encoder contains 12 layers, 12 attention heads, and a feed-forward hidden dimension of 3072. The classification head is removed, and the encoder outputs the global token $\mathbf{f}_{cls}$ and the patch-token sequence $\mathbf{F}_{patch}$ used by the subsequent pose, mask, and illumination branches.

This design is inspired by FastPose-ViT, which demonstrates that a ViT backbone can support efficient spacecraft pose estimation from cropped target images and can be coupled with a geometric recovery strategy for geometrically consistent pose prediction \cite{ancey2026fastposevit}. However, FastPose-ViT mainly relies on globally aggregated transformer representations and does not explicitly address the severe illumination variation, specular reflection, partial occlusion, and spatial reliability inconsistency commonly encountered in real orbital imagery. As a result, corrupted or unreliable local patches may still dominate the global representation, leading to unstable pose regression under challenging lighting conditions.

PAID-ViT follows this efficient crop-based transformer formulation but extends it in three aspects. First, instead of using transformer features directly for pose regression, IDTD decomposes tokens into structure and illumination streams to reduce sensitivity to lighting and preserve pose-relevant geometric information. Second, PAUP is incorporated into IDTD to estimate patch-level uncertainty and adaptively down-weight unreliable structural regions before pose regression. Third, SPM and the coarse mask branch provide additional training-side robustness and foreground-aware structural supervision, improving resistance to occlusion, highlight corruption, and background interference. Therefore, the ViT backbone serves as the shared token representation layer, while the proposed modules adapt the FastPose-ViT-style pose pipeline to illumination-challenging and reliability-varying spacecraft imagery.

\newpage
\subsection{Precision-Aware Intrinsic Disentangled Token Decomposition}
\label{subsec:idtd}
\begin{figure*}[!t]
\centerline{\includegraphics[width=0.88\textwidth]{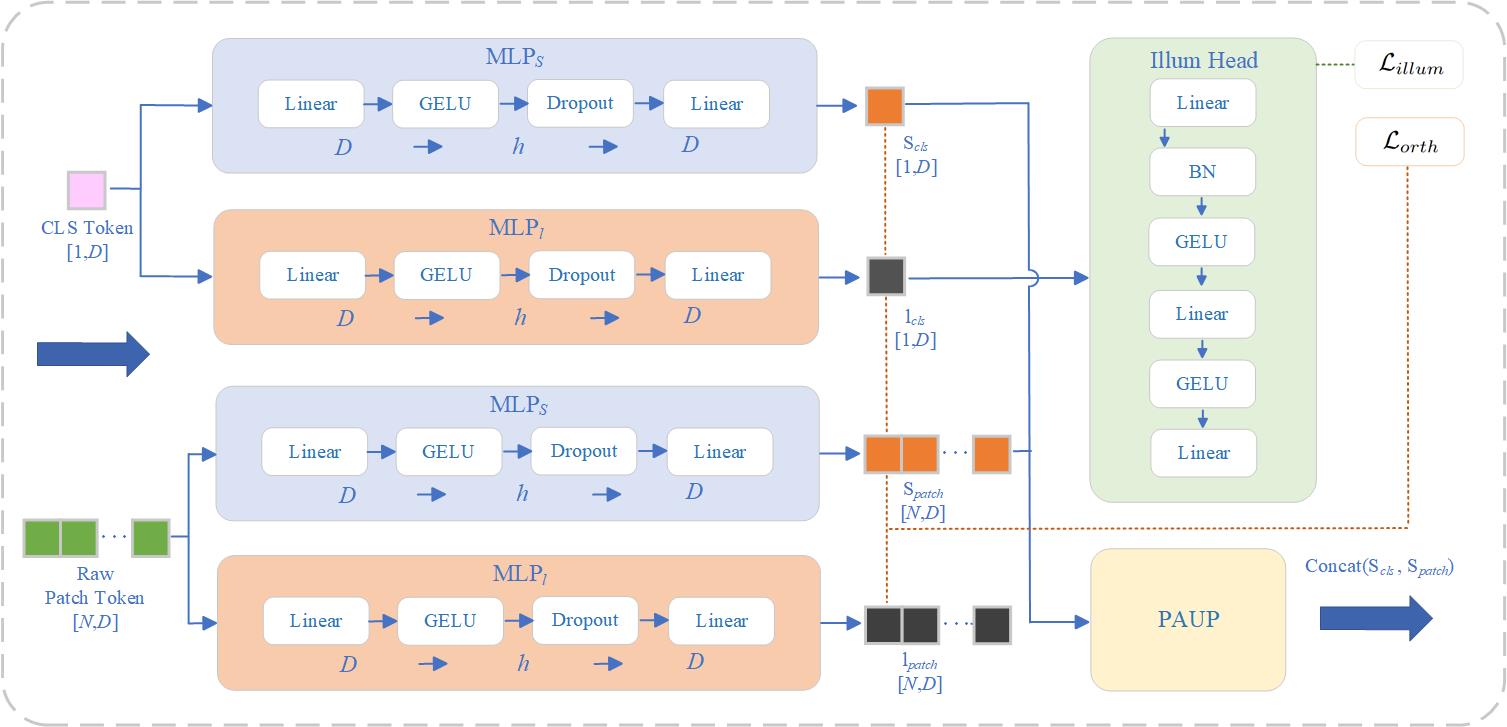}}
\caption{Illustration of the precision-aware IDTD module. The notation follows the figure: $\mathbf{S}_{cls}$ and $\mathbf{S}_{patch}$ denote structural tokens, $\mathbf{l}_{cls}$ and $\mathbf{l}_{patch}$ denote illumination tokens, and PAUP produces $\operatorname{Concat}(\mathbf{S}_{cls},\mathbf{S}_{patch})$ for the pose head.}
\label{fig:idtd_module}
\end{figure*}

\normalcolor
IDTD is designed as a pose-oriented intrinsic token decomposition module with an internal reliability estimation path. Its goal is not generic feature separation, but illumination-robust geometric representation learning under spatially varying patch quality. Consistent with Fig.~\ref{fig:idtd_module}, the CLS token and raw patch tokens are decomposed by $\mathrm{MLP}_{s}$ and $\mathrm{MLP}_{l}$ into structural tokens $(\mathbf{S}_{cls},\mathbf{S}_{patch})$ and illumination tokens $(\mathbf{l}_{cls},\mathbf{l}_{patch})$:
\begin{equation}
\mathbf{S} = \mathrm{MLP}_{s}(\mathbf{x}), \qquad
\mathbf{l} = \mathrm{MLP}_{l}(\mathbf{x}).
\label{eq:idtd}
\end{equation}
Fig.~\ref{fig:idtd_module} illustrates the internal structure of IDTD. The module receives CLS and patch tokens from the ViT backbone, separates them into structure and illumination streams, predicts patch-level reliability in the structure stream, and outputs $\operatorname{Concat}(\mathbf{S}_{cls},\mathbf{S}_{patch})$ as the reliability-weighted descriptor for the pose head. In the figures, the batch dimension is omitted for readability, and $D$ corresponds to the transformer embedding dimension $C$ used in the text.

The structure stream is responsible for object layout, panel arrangement, and silhouette geometry, while the illumination stream captures photometric factors. The illumination stream is supervised by an auxiliary head to encode attributes derived from augmentation parameters:
\begin{equation}
\hat{\mathbf{a}} = \Phi_{illum}(\mathbf{l}_{cls}) \in \mathbb{R}^{K},
\label{eq:illum}
\end{equation}
where $K$ denotes the number of illumination-related targets. To reduce information leakage between the two streams, an orthogonality regularization term is applied to normalized structure and illumination embeddings:
\begin{equation}
\mathcal{L}_{orth}
= \frac{1}{M}\sum_{i=1}^{M}
\left|\left\langle
\frac{\mathbf{S}_{i}}{\|\mathbf{S}_{i}\|_{2}},
\frac{\mathbf{l}_{i}}{\|\mathbf{l}_{i}\|_{2}}
\right\rangle\right|,
\label{eq:orth}
\end{equation}
where $M$ is the number of embeddings used in the computation. This regularization does not claim complete semantic independence; it encourages illumination-sensitive variation to be less available to the pose regressor.

PAUP is implemented inside IDTD as the reliability-aware aggregation path of the structure stream. Let $\mathbf{q}$ be the query generated from $\mathbf{S}_{cls}$, and let $\mathbf{k}_{j}$ and $\mathbf{v}_{j}$ be the key and value of the $j$-th element in $\mathbf{S}_{patch}$. A patch uncertainty head in IDTD predicts $\log\sigma_{j}^{2}$ from the corresponding structure patch. The uncertainty-adjusted attention weight is computed as
\begin{equation}
\alpha_{j} =
\mathrm{softmax}_{j}\left(
\frac{\mathbf{q}^{\top}\mathbf{k}_{j}}{\sqrt{d}}
- \frac{1}{2}\log\sigma_{j}^{2}
\right).
\label{eq:paup}
\end{equation}
The second term can be interpreted as an inverse-uncertainty correction. For the same query-key affinity, a patch with larger predicted variance receives a smaller attention weight. The pooled structural patch descriptor is denoted by $\mathbf{S}_{patch}^{paup}=\sum_{j=1}^{N}\alpha_{j}\mathbf{v}_{j}$. To match the figure notation, the final IDTD pose descriptor is written as
\begin{equation}
\mathbf{z}_{pose}
= \operatorname{Concat}(\mathbf{S}_{cls},\mathbf{S}_{patch}^{paup}).
\label{eq:idtd_descriptor}
\end{equation}
Thus, IDTD outputs not only decomposed structure and illumination tokens, but also the reliability-weighted descriptor $\mathbf{z}_{pose}$ for pose regression. Shadows, highlights, and contaminated background patches are prevented from dominating the descriptor before it reaches the pose head.

SPM is applied to the IDTD uncertainty path as a training-side regularization strategy. During training, randomly selected patches receive an uncertainty boost:
\begin{equation}
\widetilde{\log\sigma_{j}^{2}} =
\log\sigma_{j}^{2} + m_{j}\beta,
\qquad m_{j}\sim\mathrm{Bernoulli}(p_{spm}),
\label{eq:spm}
\end{equation}
where $m_{j}\in\{0,1\}$ is sampled with probability $p_{spm}$ and $\beta$ is the uncertainty boost. This operation makes IDTD less dependent on a fixed set of local patches and improves robustness when discriminative cues are partially corrupted or absent.

\subsection{Intermediate Pose Head}
\label{subsec:pose_head}
\begin{figure*}[!t]
\centerline{\includegraphics[width=0.88\textwidth]{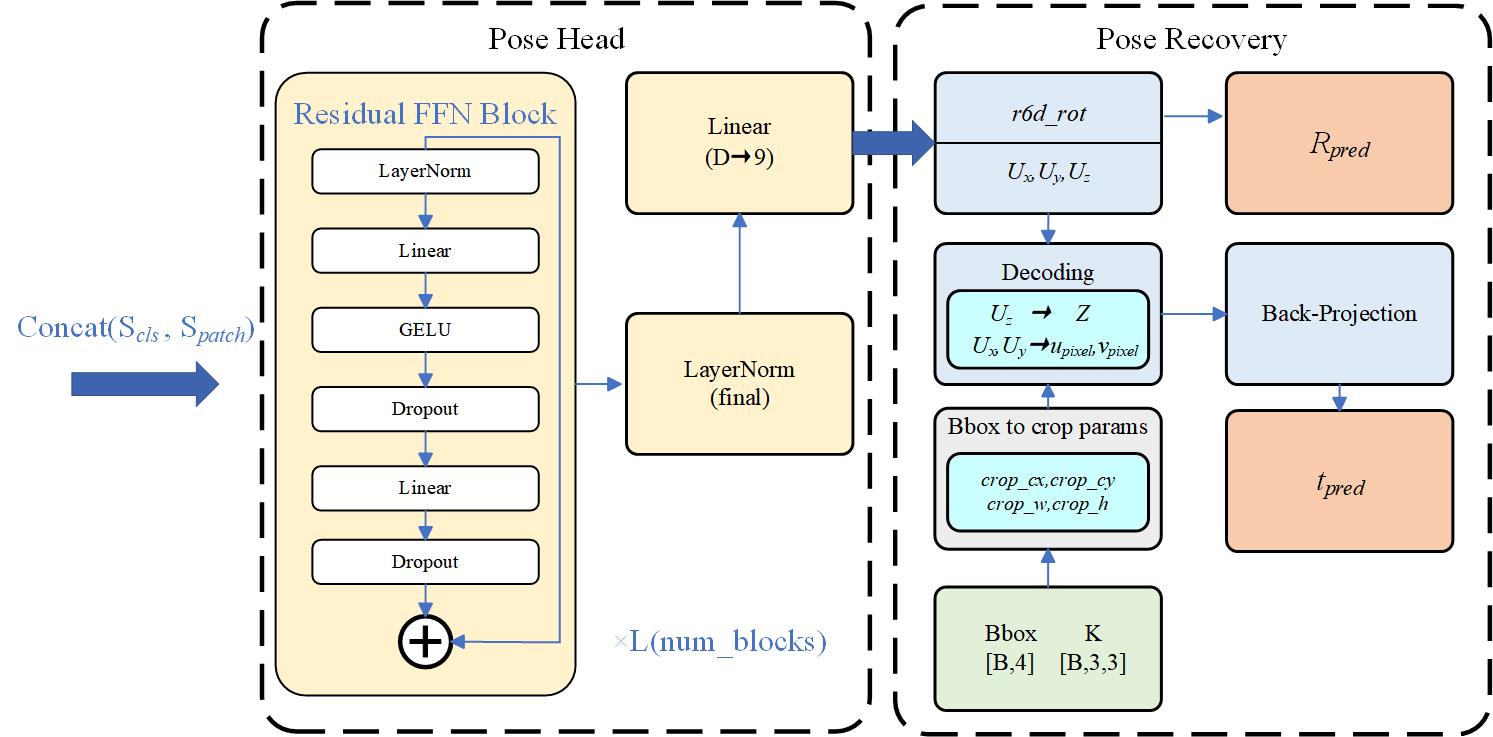}}
\caption{Illustration of the Pose Head and Pose Recovery modules. The Pose Head regresses $\mathrm{Ux,Uy,Uz}$ and $\mathrm{r6d\_rot}$ from $\operatorname{Concat}(\mathbf{S}_{cls},\mathbf{S}_{patch})$, and Pose Recovery maps them with bbox crop parameters and $\mathbf{K}$ to $\mathbf{R}_{pred}$ and $\mathbf{t}_{pred}$.}
\label{fig:pose_head_recovery}
\end{figure*}

The pose head receives the reliability-weighted IDTD descriptor $\mathbf{z}_{pose}$, shown as $\operatorname{Concat}(\mathbf{S}_{cls},\mathbf{S}_{patch})$ in Fig.~\ref{fig:pose_head_recovery}, and predicts compact intermediate variables instead of directly regressing a full pose vector:
\begin{equation}
\hat{\mathbf{y}}_{pose}
= \Phi_{pose}(\mathbf{z}_{pose})
= [\widehat{\mathrm{UxUyUz}}, \widehat{\mathrm{r6d\_rot}}]
\in \mathbb{R}^{9}.
\label{eq:pose_intermediate}
\end{equation}
Here, $\widehat{\mathrm{UxUyUz}}=[\hat{U}_{x},\hat{U}_{y},\hat{U}_{z}]$ follows the figure name for normalized crop coordinates and log-depth, and $\widehat{\mathrm{r6d\_rot}}$ denotes the continuous 6D rotation representation. The head itself is a compact pre-layer-normalized residual feed-forward regressor; reliability modeling has already been performed inside IDTD. This separation keeps the pose head focused on geometric intermediate regression while leaving patch-quality estimation to the decomposition module.

Fig.~\ref{fig:pose_head_recovery} summarizes the main pose-estimation pathway after IDTD. The pose head maps the reliability-weighted descriptor to $\mathrm{Ux,Uy,Uz}$ and $\mathrm{r6d\_rot}$, while the subsequent Pose Recovery stage converts these intermediate variables into $\mathbf{R}_{pred}$ and $\mathbf{t}_{pred}$ using bbox-derived crop parameters and camera intrinsics $\mathbf{K}$.

\vspace{-0.45\baselineskip}
\subsection{Coarse Mask Branch}
\label{subsec:mask_branch}
\begin{figure*}[!t]
\centerline{\includegraphics[width=0.88\textwidth]{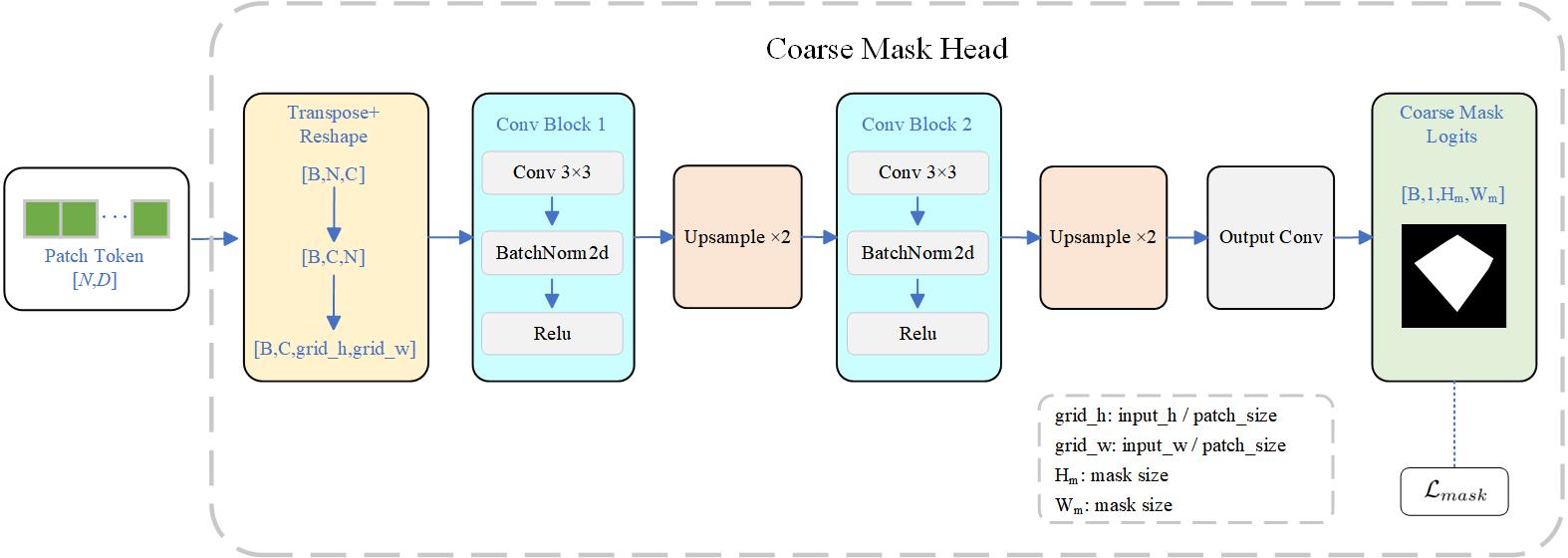}}
\caption{Illustration of the Coarse Mask Head. Raw Patch Token is transposed and reshaped into $[B,C,grid\_h,grid\_w]$, decoded by Conv Block 1/2 and upsampling layers, and supervised through Coarse Mask Logits $[B,1,H_m,W_m]$.}
\label{fig:mask_head}
\end{figure*}

The coarse mask branch provides geometry-stable auxiliary supervision. Consistent with Fig.~\ref{fig:mask_head}, it takes raw Patch Token as input and predicts Coarse Mask Logits:
\begin{equation}
\hat{\mathbf{M}}_{logit}
\in \mathbb{R}^{B \times 1 \times H_{m} \times W_{m}},
\quad H_{m}=W_{m}=56.
\label{eq:mask}
\end{equation}
Fig.~\ref{fig:mask_head} shows the mask-head design. The raw patch sequence is transposed and reshaped from $[B,N,C]$ to a spatial feature map $[B,C,grid\_h,grid\_w]$, processed by two convolutional blocks and two $\times2$ upsampling stages, and decoded by an output convolution. The branch is attached to raw patch tokens rather than IDTD structure tokens because foreground extent and silhouette boundaries remain meaningful under illumination changes and should not be overly constrained by the structure--illumination separation. This supervision encourages the shared backbone to preserve object-boundary information that is useful for pose estimation.

\subsection{Geometric Pose Recovery}
\label{subsec:recovery}
The recovery module converts learned intermediate variables into the final camera-frame pose without learnable parameters. The predicted 6D rotation vector is mapped to a valid rotation matrix by Gram--Schmidt orthogonalization:
\begin{equation}
\mathbf{R}_{pred} = \Psi_{6d}(\widehat{\mathrm{r6d\_rot}}).
\label{eq:rot_recovery}
\end{equation}
The metric depth is recovered from the logarithmic depth variable:
\begin{equation}
Z=Z_{ref}\exp(\hat{U}_{z}).
\label{eq:depth}
\end{equation}
The normalized crop coordinates $\hat{U}_{x}$ and $\hat{U}_{y}$ are decoded into image coordinates $(u_{pixel},v_{pixel})$ using bbox-derived crop parameters $(crop\_cx,crop\_cy,crop\_w,crop\_h)$. The final translation $\mathbf{t}_{pred}$ is obtained through camera back-projection with the intrinsic matrix $\mathbf{K}$. This recovery design follows the crop-based intermediate-variable formulation used by FastPose-ViT-style pipelines, while PAID-ViT modifies the perception stage by feeding the recovery module with an illumination-disentangled and reliability-weighted descriptor. Thus, PAID-ViT learns compact perceptual quantities while the final pose remains consistent with projective camera geometry.

\subsection{Training Objective}
\label{sec:loss}
The proposed network is trained in an end-to-end manner using a multi-task objective derived from the outputs of the pose branch, the mask branch, and the illumination branch. The total loss is written as
\begin{align}
\mathcal{L}
&= \mathcal{L}_{rot}
+ \mathcal{L}_{trans}
+ \mathcal{L}_{mid}
+ \mathcal{L}_{mask} \nonumber \\
&\quad
+ \mathcal{L}_{illum} \nonumber \\
&\quad
+ \mathcal{L}_{orth}
+ \mathcal{L}_{ppc}
+ \mathcal{L}_{se2}.
\label{eq:total_loss}
\end{align}
where $\mathcal{L}_{ppc}$ and $\mathcal{L}_{se2}$ are optional consistency regularizers and can be omitted in ablation studies without changing the core architecture.

The first two terms supervise the recovered physical pose after geometric recovery. The rotation loss is derived from the relative rotation on $SO(3)$:
\begin{equation}
\mathcal{L}_{rot}
= \arccos \left(
\frac{\operatorname{tr}(\hat{\mathbf{R}}\mathbf{R}^{\top})-1}{2}
\right),
\label{eq:rot_loss}
\end{equation}
where $\mathbf{R}_{pred}$ and $\mathbf{R}$ denote the predicted and ground-truth camera-frame rotations, respectively. This geodesic form directly measures angular misalignment and is more appropriate for rotation matrices than an element-wise Euclidean loss.

The translation loss is a robust regression term on the recovered camera-frame translation:
\begin{equation}
\mathcal{L}_{trans}
= \frac{1}{3}\sum_{j=1}^{3}\rho_{\beta}
\left(t_{pred,j}-t_{j}\right),
\label{eq:trans_loss}
\end{equation}
where $\rho_{\beta}(\cdot)$ denotes the Smooth-$L_{1}$ penalty. This term constrains the final metric translation and reduces the influence of occasional large regression errors.

In addition to supervising the recovered pose, PAID-ViT supervises the FastPose-style intermediate variables before recovery. Following Fig.~\ref{fig:pose_head_recovery}, the intermediate vector is defined as $\mathbf{y}=[\mathrm{Ux,Uy,Uz},\mathrm{r6d\_rot}^{\top}]^{\top}$, where $U_x$ and $U_y$ are normalized crop-centered coordinates, $U_z=\log(Z/Z_{ref})$ encodes relative depth, and $\mathrm{r6d\_rot}$ is converted to a rotation matrix through Gram--Schmidt orthogonalization. These targets are obtained from the ground-truth pose and crop metadata by the parameter-free recovery model, so they provide geometrically consistent supervision without additional annotation.

When standard intermediate regression is used, the loss is
\begin{equation}
\mathcal{L}_{mid}
= \frac{1}{3}\sum_{j=1}^{3}\rho_{\beta}
\left(\hat{U}_{j}-U_{j}\right)
+ \frac{1}{6}\sum_{j=1}^{6}\rho_{\beta}
\left(\hat{r}_{j}-r_{j}\right).
\label{eq:mid_smooth}
\end{equation}
This term stabilizes optimization by giving the pose head direct supervision on the variables it actually predicts, instead of relying only on the final recovered rotation and translation.

When heteroscedastic intermediate regression is enabled, the pose head additionally predicts $\hat{\mathbf{v}}=\log\boldsymbol{\sigma}^{2}_{pose}\in\mathbb{R}^{9}$. The intermediate loss is replaced by a Gaussian negative log-likelihood (NLL):
\begin{align}
\mathcal{L}_{mid}
&= \mathcal{L}_{nll}, \nonumber\\
\mathcal{L}_{nll}
&= \frac{1}{9}\sum_{i=1}^{9}
\left[
\frac{1}{2}\exp(-\hat{v}_{i})(y_{i}-\hat{y}_{i})^{2}
+ \frac{1}{2}\hat{v}_{i}
\right],
\label{eq:nll}
\end{align}
where $\hat{v}_{i}$ is the predicted log variance of the $i$-th intermediate variable. The quadratic residual is weighted by the inverse variance $\exp(-\hat{v}_{i})$, while the log-variance term prevents the network from trivially increasing uncertainty. This loss is used to model aleatoric ambiguity in difficult pose components. The patch uncertainty used by PAUP inside IDTD is not directly supervised by labels; it is learned through the pose objective because patches assigned larger uncertainty contribute less to the reliability-weighted IDTD descriptor.

The coarse mask branch is supervised by foreground masks resized to $56\times56$. Its loss combines binary cross-entropy with Dice overlap:
\begin{align}
\mathcal{L}_{mask}
&= \operatorname{BCE}(\hat{\mathbf{M}}_{logit},\mathbf{M})
 + \mathcal{L}_{dice}, \nonumber\\
\mathcal{L}_{dice}
&= 1-
\frac{2\sum \sigma(\hat{\mathbf{M}}_{logit})\mathbf{M}+\epsilon}
{\sum \sigma(\hat{\mathbf{M}}_{logit})+\sum \mathbf{M}+\epsilon}.
\label{eq:mask_loss}
\end{align}
where $\hat{\mathbf{M}}_{logit}$ denotes Coarse Mask Logits and $\mathbf{M}$ denotes the binary foreground mask. The BCE term provides pixel-level classification supervision, whereas the Dice term emphasizes foreground overlap and is useful when the spacecraft occupies only a limited portion of the crop. This auxiliary loss encourages the backbone to retain silhouette and boundary cues that are stable under illumination changes.

The illumination loss is generated from the IDTD illumination stream. Let $\hat{\mathbf{a}}\in\mathbb{R}^{K}$ be the output of the illumination head and $\mathbf{a}\in\mathbb{R}^{K}$ be the scalar target vector converted from photometric augmentation parameters. We define
\begin{equation}
\mathcal{L}_{illum}
= \frac{1}{K}\sum_{k=1}^{K}
\rho_{\beta}\left(\hat{a}_{k}-a_{k}\right).
\label{eq:illum_loss}
\end{equation}
This term gives the illumination tokens an explicit photometric prediction task. As a result, illumination-sensitive variation is encouraged to be represented by the illumination stream rather than leaking into the structure tokens used by the pose head.

The orthogonality loss $\mathcal{L}_{orth}$ is defined in (\ref{eq:orth}) and is applied to normalized structure and illumination embeddings. It penalizes high cosine similarity between the two streams, reducing redundant information sharing and strengthening the intended division between geometric and photometric factors.

Finally, projected pose consistency (PPC), frequency-domain perturbation/mixing (FPM), and $SE(2)$ consistency are used as training-side robustness components. $\mathcal{L}_{ppc}$ enforces consistency between weakly and strongly augmented views, while $\mathcal{L}_{se2}$ encourages the prediction to remain consistent under in-plane image transformations after accounting for the corresponding crop geometry. FPM mixes spectral amplitude information in the data pipeline to perturb image appearance without changing the pose labels. These losses and perturbations are not required for inference; they regularize the representation during training and can be used as optional robustness components.

\section{Experiments}
\label{sec:experiments}
This section evaluates PAID-ViT from four aspects: experimental settings, official and physical pose metrics, comparison with representative spacecraft pose-estimation methods, and ablation analysis. The evaluation follows the SPEED+ protocol because this benchmark explicitly separates synthetic validation images from lightbox and sunlamp hardware-in-the-loop domains, making it suitable for analyzing illumination variation and domain shift.

\begin{figure*}[!t]
\centerline{\includegraphics[width=0.90\textwidth]{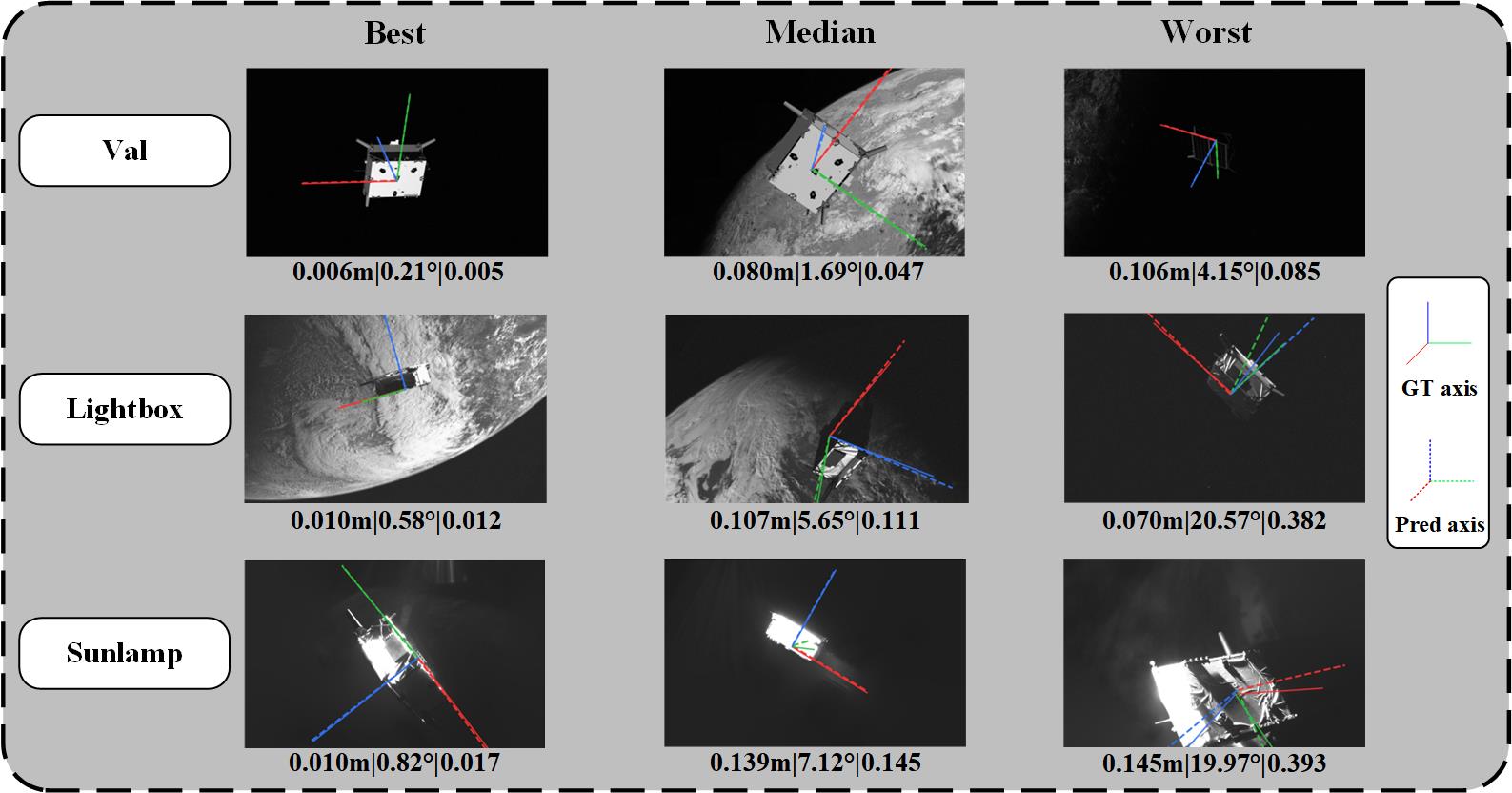}}
\caption{Representative qualitative pose-estimation examples on SPEED+ V2. The rows correspond to validation, lightbox, and sunlamp domains, and the columns show best, median, and worst cases. Solid axes indicate ground truth and dashed axes indicate predictions. Numeric overlays, when shown, follow the $E_R/E_t$ format, where $E_R$ is in degrees and $E_t$ is in meters.}
\label{fig:speedplusv2_cases}
\end{figure*}

\begin{table*}[!t]
\caption{Comparison with representative methods on SPEED+ / SPEED+ V2. $S_{pose}$ is the unitless official score, $E_t$ is measured in meters, and $E_R$ is measured in degrees. Lower values are better. ``--'' denotes metrics not reported in the cited source or official record. Bold values highlight the best physical error among direct no-PnP methods.}
\label{tab:sota_comparison}
\centering
\footnotesize
\begin{tabular}{L{0.19\textwidth}c c c c c c c c}
\toprule
Method & Year & PnP & Lightbox $S_{pose}$ & Sunlamp $S_{pose}$ & Lightbox $E_t$ & Lightbox $E_R$ & Sunlamp $E_t$ & Sunlamp $E_R$ \\
\midrule
\multicolumn{9}{l}{\textit{Solver-assisted or hybrid pose-estimation methods}}\\
EagerNet \cite{eagernet2023} & 2023 & Yes & 0.039 & 0.059 & -- & -- & -- & -- \\
SPNv2 ODR \cite{park2023spnv2} & 2023 & Yes & 0.122 & 0.197 & 0.150 & 5.577 & 0.161 & 9.788 \\
PVSPE \cite{pvspe2024} & 2024 & -- & 0.101 & 0.178 & -- & 4.810 & -- & 8.940 \\
PVSAR \cite{pvsar2024} & 2024 & Yes & 0.076 & 0.112 & -- & -- & -- & -- \\
UAKD \cite{uakd2025} & 2025 & Yes & 0.248 & 0.360 & 0.288 & 11.420 & 0.373 & 17.030 \\
UF-SPE \cite{ufspe2025} & 2025 & Yes & 0.090 & 0.120 & 0.112 & 3.970 & 0.110 & 5.690 \\
\midrule
\multicolumn{9}{l}{\textit{Direct no-PnP pose-regression methods}}\\
FA-VAE \cite{favae2024} & 2024 & No & 0.114 & 0.118 & -- & -- & -- & -- \\
FastPose-ViT \cite{ancey2026fastposevit} & 2026 & No & -- & -- & 0.270 & \textbf{12.261} & 0.400 & 18.277 \\
Cov2Pose \cite{cov2pose2026} & 2026 & No & 0.310 & 0.574 & 0.300 & 15.096 & 0.430 & 29.128 \\
PAID-ViT (Ours) & 2026 & No & 0.237 & 0.230 & \textbf{0.135} & 12.291 & \textbf{0.151} & \textbf{11.760} \\
\bottomrule
\end{tabular}
\end{table*}

\subsection{Experimental Settings}
\label{subsec:experimental_settings}
Experiments are conducted on the SPEED+ benchmark \cite{park2022speedplus}. In this article, SPEED+ V2 denotes the SPEED+ validation/lightbox/sunlamp evaluation configuration used in our experiment records and qualitative visualizations. The dataset contains synthetic training data and three evaluation domains: validation, lightbox, and sunlamp. The validation split contains 11994 images, while the lightbox and sunlamp domains contain 6740 and 2791 hardware-in-the-loop images, respectively. The lightbox subset provides controlled laboratory illumination, whereas the sunlamp subset introduces stronger shadows, specular highlights, and lower-contrast target regions. These two domains are therefore used as the main cross-domain tests.

The implementation is based on a PyTorch/CUDA workflow with a timm ViT backbone. Unless otherwise stated, ablation results are reported with exponential moving average (EMA) weights.

\subsection{Evaluation Metrics}
\label{subsec:evaluation_metrics}
Following the official SPEED+ scoring rule \cite{esa2021specscoring}, we use the normalized translation score, the rotation score, and their sum:
\begin{equation}
S_t = \frac{\left\|\hat{\mathbf{t}}-\mathbf{t}\right\|_2}{\left\|\mathbf{t}\right\|_2},
\label{eq:speed_translation_score}
\end{equation}
\begin{equation}
S_R = 2\arccos\left(\left|\langle \hat{\mathbf{q}}, \mathbf{q}\rangle\right|\right),
\label{eq:speed_rotation_score}
\end{equation}
\begin{equation}
S_{pose}=S_t+S_R,
\label{eq:speed_pose_score}
\end{equation}
where $\hat{\mathbf{t}}$ and $\mathbf{t}$ are the predicted and ground-truth translations, and $\hat{\mathbf{q}}$ and $\mathbf{q}$ are the predicted and ground-truth unit quaternions. Lower values indicate better performance.

In addition to the official normalized scores, we report physical translation and rotation errors:
\begin{equation}
E_t=\left\|\hat{\mathbf{t}}-\mathbf{t}\right\|_2,
\label{eq:physical_translation_error}
\end{equation}
\begin{equation}
E_R=\frac{180}{\pi}\arccos\left(\frac{\mathrm{tr}\left(\hat{\mathbf{R}}^{T}\mathbf{R}\right)-1}{2}\right).
\label{eq:physical_rotation_error}
\end{equation}

Fig.~\ref{fig:speedplusv2_cases} illustrates representative SPEED+ V2 qualitative cases across the validation, lightbox, and sunlamp domains. The rows correspond to validation, lightbox, and sunlamp images, and the columns show best, median, and worst cases. Solid axes denote ground-truth pose axes, whereas dashed axes denote predicted pose axes. The examples indicate that illumination and domain shift can substantially change pose difficulty even when the spacecraft geometry is unchanged.

\begin{table*}[!t]
\caption{Cumulative ablation results on hardware-in-the-loop domains with EMA weights. $E_t$ is measured in meters and $E_R$ in degrees.}
\label{tab:ablation_progression}
\centering
\begin{tabular}{L{0.18\textwidth}L{0.34\textwidth}cc cc}
\toprule
Stage & Enabled components & \multicolumn{2}{c}{Lightbox} & \multicolumn{2}{c}{Sunlamp} \\
\cmidrule(lr){3-4}\cmidrule(lr){5-6}
 & & $E_R$ & $E_t$ & $E_R$ & $E_t$ \\
\midrule
G0 & ViT + CLS pose head & 18.108 & 0.193 & 18.947 & 0.203 \\
G1 & G0 + mask branch & 15.193 & 0.182 & 15.782 & 0.199 \\
G2 & G1 + IDTD/PAUP/SPM/NLL & 13.452 & 0.167 & 13.330 & 0.180 \\
G3 & G2 + PPC/FPM/$SE(2)$/Illum & 12.291 & 0.135 & 11.760 & 0.151 \\
\bottomrule
\end{tabular}
\end{table*}

\subsection{Comparison With Representative Methods}
\label{subsec:sota_comparison}
Table~\ref{tab:sota_comparison} summarizes representative SPEED+ and SPEED+ V2 results from recent spacecraft pose-estimation studies and official SPEC2021 records. The table intentionally includes solver-assisted, hybrid, and direct-regression methods, so the results should be interpreted by formulation rather than as a fully controlled architecture-only comparison. External values are taken from the cited public papers, supplementary materials, or official records; missing entries are marked by ``--'' when the corresponding source does not report that metric. Solver-assisted or hybrid methods, such as SPNv2 ODR, PVSAR, UAKD, and UF-SPE, use explicit geometric intermediates, online refinement, or PnP-style recovery after perception. These additional constraints explain their strong official scores and, for several entries, lower rotation errors. For example, UF-SPE and SPNv2 ODR benefit from solver-side geometric refinement, while UAKD emphasizes compact uncertainty-aware distillation and therefore remains competitive in lightbox rotation. The remaining gap against these methods is mainly related to the absence of geometry-constrained post-processing in PAID-ViT. In return, PAID-ViT does not rely on keypoint handoff or a PnP solver, which simplifies the inference pipeline and may reduce sensitivity to unstable intermediate correspondences under highlights, shadows, or partial foreground degradation.

Among direct no-PnP pose-regression methods, FastPose-ViT is the closest reference because it also uses a crop-based ViT representation, and its values are taken from the public FastPose-ViT paper. Cov2Pose is included as a recent direct-regression reference using the reported SPEED+ / SPEED+ V2 protocol values. FA-VAE reports official pose scores but does not provide physical translation and rotation errors in Table~\ref{tab:sota_comparison}, so it is used as a score-level reference rather than an error-level comparison. Based on the physical errors in Table~\ref{tab:sota_comparison}, PAID-ViT obtains $S_{pose}$ values of 0.237 on lightbox and 0.230 on sunlamp. PAID-ViT reduces the lightbox and sunlamp translation errors of FastPose-ViT from 0.270/0.400 m to 0.135/0.151 m, corresponding to relative reductions of 50.0\% and 62.3\%, respectively. FastPose-ViT reports a slightly lower lightbox rotation error, which is consistent with its globally aggregated transformer descriptor being effective under controlled illumination. In contrast, PAID-ViT obtains a lower rotation error on the more difficult sunlamp domain, reducing the reported value from 18.277$^\circ$ to 11.760$^\circ$. These results suggest that IDTD and precision-aware token aggregation are useful for reducing the influence of illumination-sensitive and unreliable local evidence before pose regression. Compared with Cov2Pose, PAID-ViT obtains lower translation and rotation errors on both hardware-in-the-loop domains, suggesting that explicit illumination-oriented token modeling and patch reliability estimation are beneficial under strong domain shift.

\raggedbottom
\subsection{Ablation Study}
\label{subsec:ablation}
The ablation study follows a cumulative design. All ablation models are trained for 90 epochs with a fixed random seed of 42. The batch size is 24 per GPU, the number of data-loading workers is 16, and the learning rate is scaled by the square-root rule with a base batch size of 32. Each comparison follows the same CLS-based pose-regression setting. As summarized in Table~\ref{tab:ablation_progression}, the table omits the synthetic validation split and focuses on the lightbox and sunlamp hardware-in-the-loop domains. The baseline G0 corresponds to a FastPose-style ViT-Base model with a CLS pose head and smooth-$L_1$ intermediate supervision. G1 adds only the coarse mask branch. G2 adds the proposed structure regularization group, including IDTD, internal PAUP, SPM, orthogonality loss, and heteroscedastic negative log-likelihood (NLL). G3 further adds training-side consistency and input perturbation, including PPC, FPM, $SE(2)$ consistency, and illumination supervision. Because the study is cumulative, the results support the effectiveness of module groups under a shared training protocol rather than proving that every subcomponent is independently optimal.

\subsubsection{Mask Branch}
The mask branch provides an auxiliary geometry-stable silhouette signal. As shown in Table~\ref{tab:mask_ablation}, adding mask supervision improves the lightbox and sunlamp rotation errors and also reduces the translation error on both hardware-in-the-loop domains. The improvement suggests that foreground-aware supervision helps the shared transformer features preserve target boundaries under domain shift.

\begingroup
\setlength{\intextsep}{3pt}
\setlength{\abovecaptionskip}{1pt}
\setlength{\belowcaptionskip}{2pt}
\begin{table}[H]
\caption{Ablation of the coarse mask branch. $E_R$ is in degrees and $E_t$ is in meters.}
\label{tab:mask_ablation}
\centering
\scriptsize
\setlength{\tabcolsep}{2pt}
\begin{tabular}{lcccc}
\toprule
Stage & Light $E_R$ & Light $E_t$ & Sun $E_R$ & Sun $E_t$ \\
\midrule
G0 & 18.108 & 0.193 & 18.947 & 0.203 \\
G1 & 15.193 & 0.182 & 15.782 & 0.199 \\
\bottomrule
\end{tabular}
\end{table}
\endgroup

\begingroup
\setlength{\intextsep}{3pt}
\setlength{\abovecaptionskip}{1pt}
\setlength{\belowcaptionskip}{2pt}
\begin{figure}[H]
\centerline{\includegraphics[width=0.90\columnwidth]{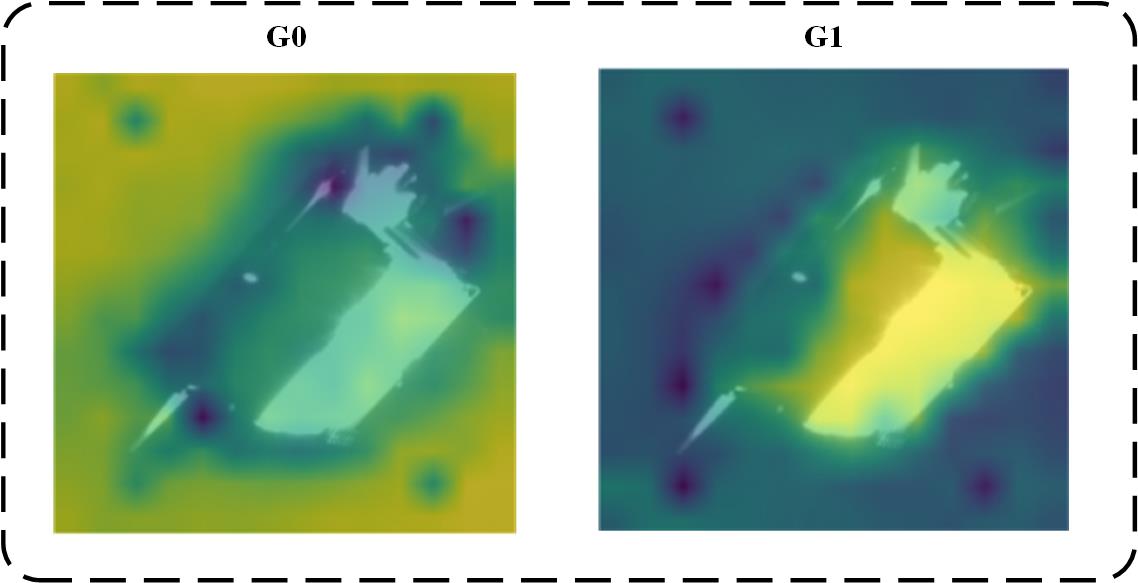}}
\caption{Grad-CAM visualization of G1 relative to G0. The responses suggest that the mask branch helps concentrate attention on spacecraft foreground regions and reduces background activation.}
\label{fig:g1_g0_gradcam}
\end{figure}
\endgroup

Fig.~\ref{fig:g1_g0_gradcam} further visualizes the effect of the mask branch using Gradient-weighted Class Activation Mapping. Compared with G0, G1 produces responses that are more concentrated on the spacecraft foreground and less dispersed over the background, supporting the foreground-constraint role indicated by Table~\ref{tab:mask_ablation}.

\subsubsection{IDTD With PAUP and SPM}
IDTD with PAUP and SPM is the central feature-space robustness component. Table~\ref{tab:idtd_ablation} shows that G2 reduces translation and rotation errors on both hardware-in-the-loop domains. This result supports the intended role of IDTD: illumination-sensitive appearance factors are encouraged to be separated from structure tokens, and unreliable local patches are down-weighted before pose aggregation.

\begingroup
\setlength{\intextsep}{3pt}
\setlength{\abovecaptionskip}{1pt}
\setlength{\belowcaptionskip}{2pt}
\begin{table}[H]
\caption{Ablation of IDTD with internal PAUP, SPM, and NLL. $E_R$ is in degrees and $E_t$ is in meters.}
\label{tab:idtd_ablation}
\centering
\scriptsize
\setlength{\tabcolsep}{2pt}
\begin{tabular}{lcccc}
\toprule
Stage & Light $E_R$ & Light $E_t$ & Sun $E_R$ & Sun $E_t$ \\
\midrule
G1 & 15.193 & 0.182 & 15.782 & 0.199 \\
G2 & 13.452 & 0.167 & 13.330 & 0.180 \\
\bottomrule
\end{tabular}
\end{table}
\endgroup

To further examine the token source used by the pose head, we conduct a diagnostic retraining experiment with three variants. As summarized in Table~\ref{tab:idtd_token_source_ablation}, T0-s uses the structural token produced by IDTD, T1-raw-z uses the raw transformer descriptor, and T3-s+l mixes the structure and illumination streams. The variants are trained for 50 epochs and evaluated with EMA weights. Compared with the raw descriptor, the structural token gives a small rotation advantage on the real lightbox and sunlamp domains. More importantly, mixing $s$ and $l$ degrades real-domain rotation, indicating that directly reintroducing illumination-sensitive information into the pose input can disturb cross-domain pose regression. Therefore, this experiment supports the proposed pose-head design without claiming complete semantic independence between the two streams.

\begingroup
\setlength{\intextsep}{3pt}
\setlength{\abovecaptionskip}{1pt}
\setlength{\belowcaptionskip}{2pt}
\begin{table}[H]
\caption{Diagnostic ablation of IDTD pose-token source. Each cell reports $E_R/E_t$, where $E_R$ is in degrees and $E_t$ is in meters.}
\label{tab:idtd_token_source_ablation}
\centering
\setlength{\tabcolsep}{2pt}
\begin{tabular}{lccc}
\toprule
Variant & Val & Lightbox & Sunlamp \\
\midrule
T0-s & 2.70 / 0.0386 & 15.47 / 0.1596 & 15.25 / 0.1755 \\
T1-raw-z & 2.70 / 0.0390 & 15.90 / 0.1634 & 15.56 / 0.1738 \\
T3-s+l & 2.69 / 0.0391 & 16.34 / 0.1618 & 16.39 / 0.1780 \\
\bottomrule
\end{tabular}
\end{table}
\endgroup

\subsubsection{Training-Side Robustness}
The final robustness group adds PPC, FPM, $SE(2)$ consistency, and illumination supervision. As shown in Table~\ref{tab:robust_ablation}, this group further improves both translation and rotation errors on lightbox and sunlamp, producing the best hardware-domain results in the cumulative ablation. These gains suggest that training-side perturbation and consistency regularization can improve robustness without changing the inference architecture.

\begingroup
\setlength{\intextsep}{3pt}
\setlength{\abovecaptionskip}{1pt}
\setlength{\belowcaptionskip}{2pt}
\begin{table}[H]
\caption{Ablation of training-side robustness regularization. $E_R$ is in degrees and $E_t$ is in meters.}
\label{tab:robust_ablation}
\centering
\scriptsize
\setlength{\tabcolsep}{2pt}
\begin{tabular}{lcccc}
\toprule
Stage & Light $E_R$ & Light $E_t$ & Sun $E_R$ & Sun $E_t$ \\
\midrule
G2 & 13.452 & 0.167 & 13.330 & 0.180 \\
G3 & 12.291 & 0.135 & 11.760 & 0.151 \\
\bottomrule
\end{tabular}
\end{table}
\endgroup

\flushbottom
\section{Conclusion}
\label{sec:conclusion}
This article presented PAID-ViT, a precision-aware illumination-disentangled Vision Transformer for monocular spacecraft 6D pose estimation. The framework combines transformer token representation, precision-aware intrinsic token decomposition, IDTD-internal uncertainty-guided structure aggregation, stochastic patch masking, coarse foreground supervision, and parameter-free geometric recovery. Experimental results on SPEED+ V2 suggest that the proposed reliability-aware representation reduces translation error and improves robustness under challenging sunlamp illumination, while the ablation study supports the complementary roles of foreground supervision, token decomposition, and training-side robustness regularization.

Several limitations remain. The current evaluation is mainly conducted on SPEED+ V2 hardware-in-the-loop domains, and further validation on real in-orbit imagery and more severe occlusion cases is still needed. PAID-ViT also does not use PnP, online refinement, or multi-frame filtering, so solver-assisted methods with explicit geometric post-processing may still achieve lower official scores in some settings. Future work will investigate stronger real-domain adaptation and the integration of lightweight geometric refinement while preserving the direct-regression inference pipeline.

\bibliographystyle{IEEEtran}
\bibliography{references}

\newpage

\makeatletter
\def\@IEEEBIOskipN{1.1\baselineskip}
\makeatother

\begin{IEEEbiography}[{\includegraphics[width=1in,height=1.25in,clip,keepaspectratio]{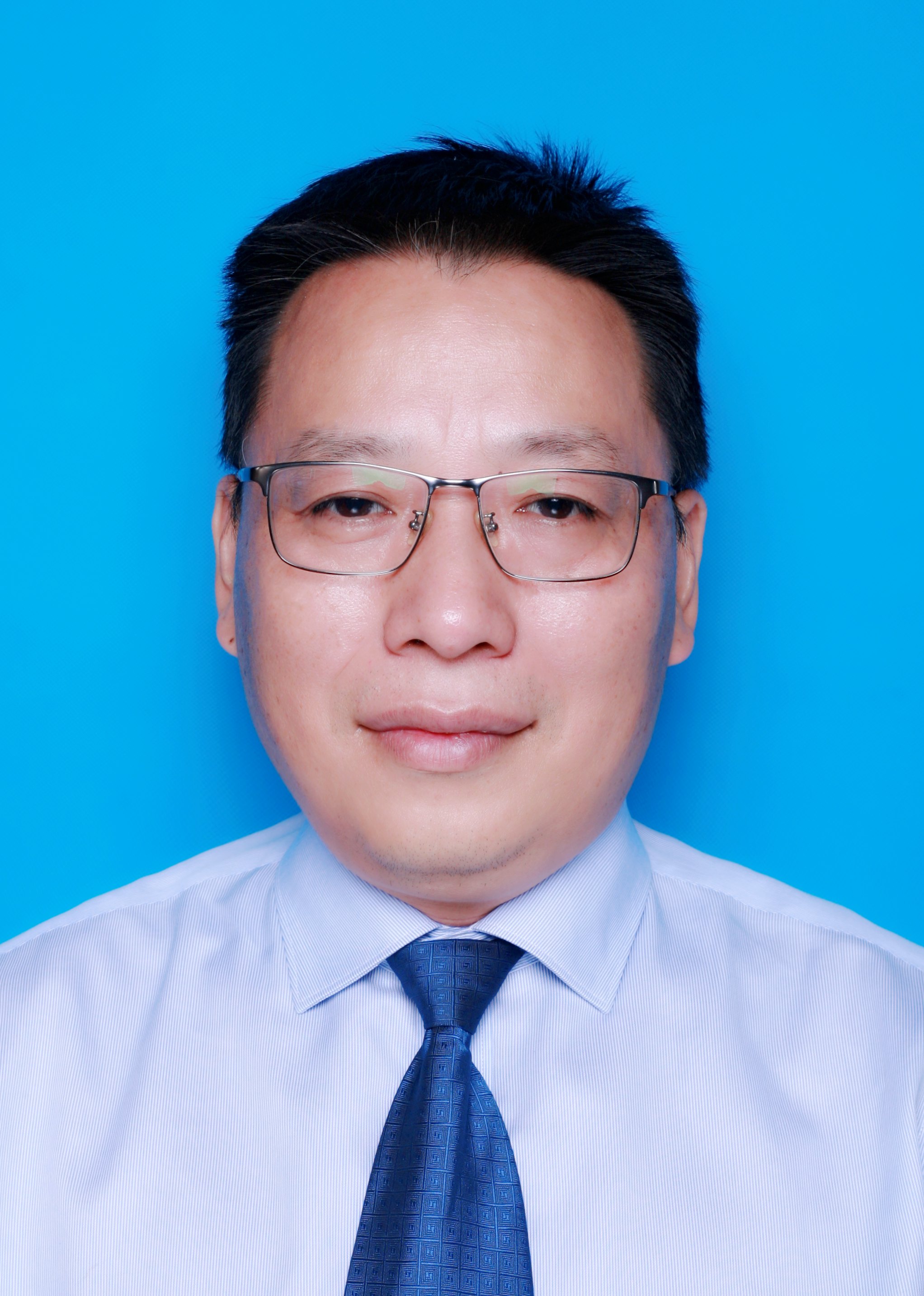}}]{Zongwu Xie}
received the B.S. degree in electrical engineering and automation from Harbin University of Science and Technology, Harbin, China, in 1996, and the M.S. and Ph.D. degrees in mechanical engineering from Harbin Institute of Technology, Harbin, China, in 2000 and 2003, respectively.
\end{IEEEbiography}

\begin{IEEEbiography}[{\includegraphics[width=1in,height=1.25in,clip,keepaspectratio]{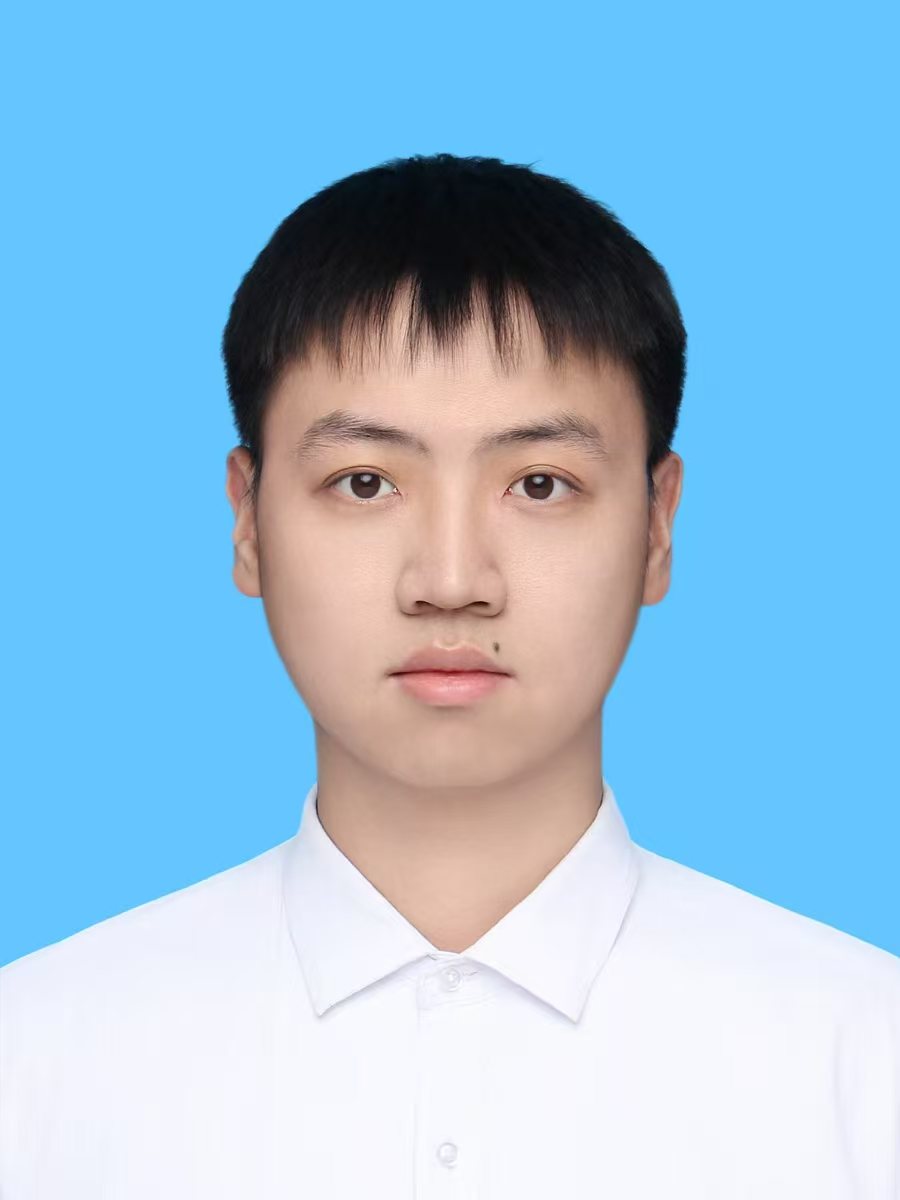}}]{Yifan Yang}
received the B.S. degree in mechanical design, manufacturing, and automation from Harbin Institute of Technology, Harbin, China, in 2025. He is currently pursuing the M.S. degree with the School of Mechatronics Engineering, Harbin Institute of Technology, as a member of the Class of 2025. His research interests include robot pose estimation and arm manipulation.
\end{IEEEbiography}

\begin{IEEEbiography}[{\includegraphics[width=1in,height=1.25in,clip,keepaspectratio]{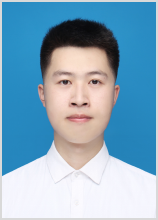}}]{Yonglong Zhang}
received the B.S. degree in mechatronic engineering from Harbin Institute of Technology, Harbin, China, in 2024. He is currently pursuing the M.S. degree with the School of Mechatronics Engineering, Harbin Institute of Technology, as a member of the Class of 2024. His research interests include robot pose estimation and arm manipulation.
\end{IEEEbiography}

\begin{IEEEbiography}[{\includegraphics[width=1in,height=1.25in,clip,keepaspectratio]{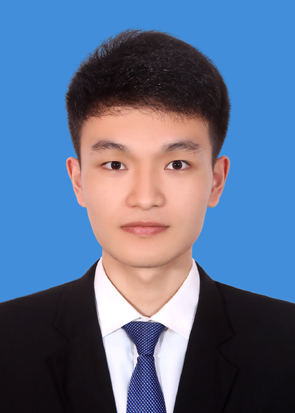}}]{Guanghu Xie}
received his Bachelor's degree in Mechatronic Engineering from Harbin Institute of Technology in 2020 and his Master's degree in Mechanical Engineering from Harbin Institute of Technology in 2022. He is currently pursuing the Ph.D. degree in mechanical engineering with Harbin Institute of Technology. His main research focus is on dexterous manipulation using robotic arms and hands.
\end{IEEEbiography}

\begin{IEEEbiography}[{\includegraphics[width=1in,height=1.25in,clip,keepaspectratio]{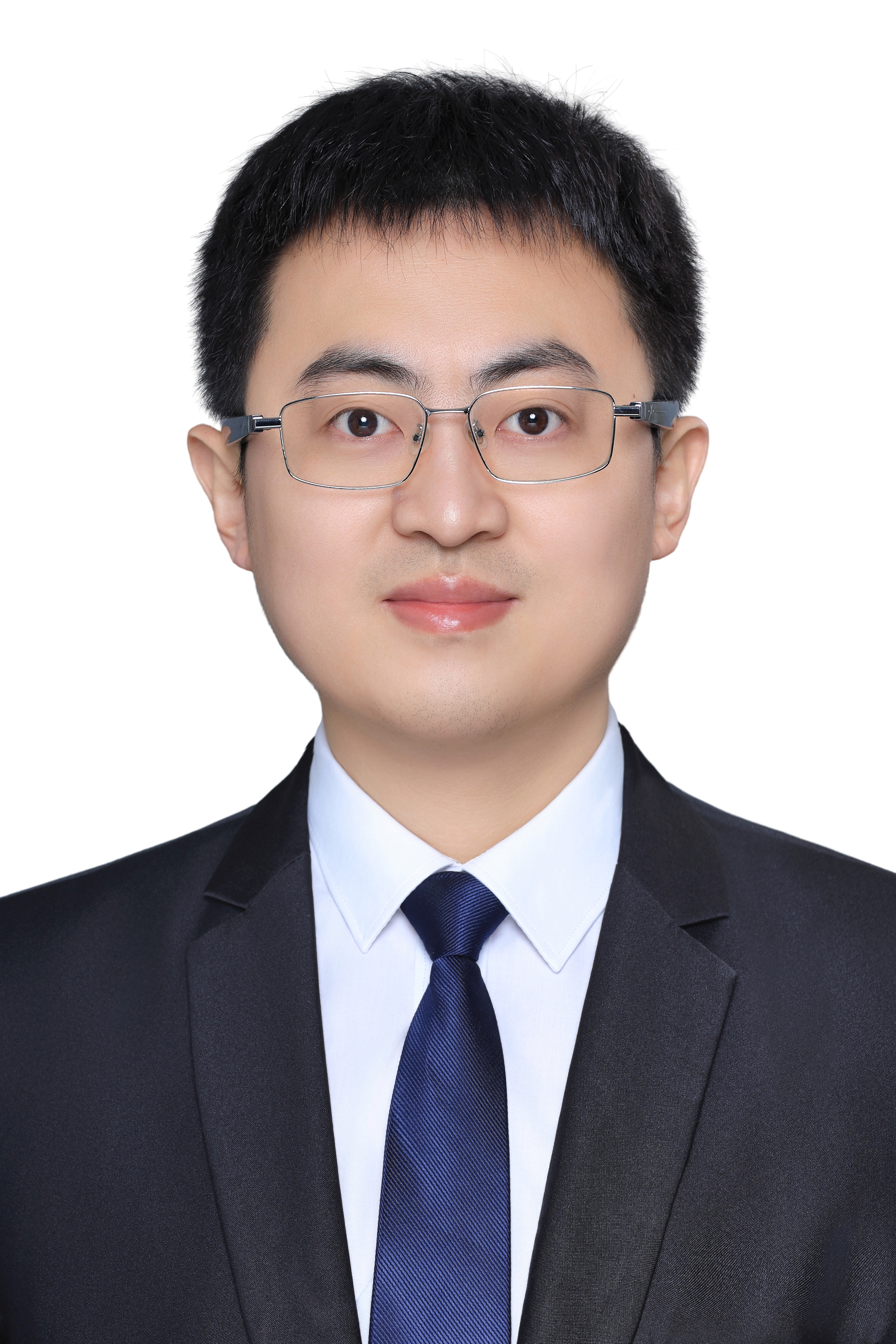}}]{Yang Liu}
was born in Handan, Hebei, China, in 1990. He received the B.S., M.S., and Ph.D. degrees from Harbin Institute of Technology, Harbin, China, in 2013, 2015, and 2020, respectively. He is currently an Associate Professor with Harbin Institute of Technology and serves as the Deputy Secretary of the Youth League Committee. His research interests include space robotics, humanoid robots, and computer vision. He has published 35 scientific papers, 16 invention patents, and two textbooks. He has been selected for the Youth Talent Support Program and honored with titles including Heilongjiang Province Outstanding Youth.
\end{IEEEbiography}

\begin{IEEEbiography}[{\includegraphics[width=1in,height=1.25in,clip,keepaspectratio]{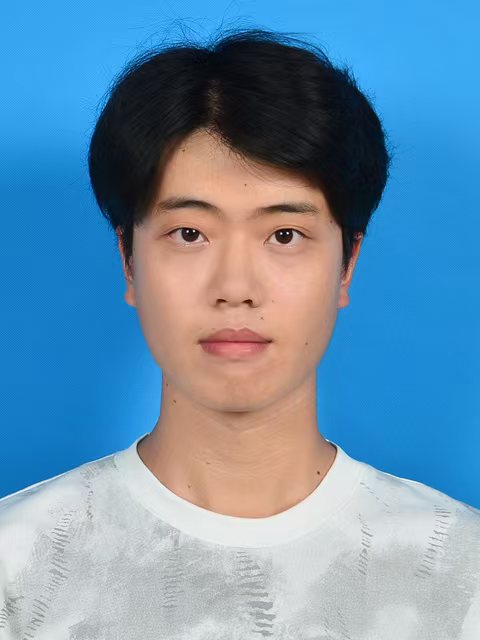}}]{Shuo Zhang}
received the B.S. degree in robotics engineering from Harbin Institute of Technology, Harbin, China, in 2025. He is currently pursuing the M.S. degree with the School of Mechatronics Engineering, Harbin Institute of Technology, as a member of the Class of 2025. His research interests include robot reinforcement learning, robotic manipulation, and robot pose estimation.
\end{IEEEbiography}

\end{document}